\documentclass[conference]{IEEEtran}
\IEEEoverridecommandlockouts

\usepackage{wrapfig}
\usepackage{scrextend}
\usepackage{epsfig}
\usepackage{setspace}
\usepackage{varwidth}
\usepackage{bbding}
\usepackage{microtype}
\usepackage{graphicx}
\usepackage{subfigure}
\usepackage{changepage}
\usepackage{multirow}
\usepackage{tabularx}
\usepackage{makecell}
\usepackage{pifont}
\usepackage{bbding}
\usepackage{tikz}
\usepackage{pifont}
\usepackage{color}
\usepackage{enumerate}
\usepackage{enumitem}
\usepackage{diagbox}
\usepackage{parskip}
\usepackage{booktabs}
\usepackage{soul}
\usepackage{pifont}
\usepackage{dashrule}
\usepackage{colortbl}
\definecolor{tred}{RGB}{128,0,32}
\definecolor{tgreen}{RGB}{34,139,34}
\definecolor{lightgray}{gray}{0.93}

\usepackage{marvosym}
\makeatletter
\renewcommand{\footnoterule}{%
  \kern-3pt
  \hrule\@width 1in % 横线的长度
  \@height 0.4pt % 横线的厚度
  \kern 2.6pt
}
\makeatother
\usepackage[symbol*]{footmisc} % 使用符号作为脚注标记
\usepackage{amsmath,amsfonts,bm}

\def\eqref#1{equation~\ref{#1}}
\def\1{\bm{1}}

\DeclareMathAlphabet{\mathsfit}{\encodingdefault}{\sfdefault}{m}{sl}
\SetMathAlphabet{\mathsfit}{bold}{\encodingdefault}{\sfdefault}{bx}{n}

\usepackage[utf8]{inputenc} % allow utf-8 input
\usepackage{xcolor}         % colors

\def\BibTeX{{\rm B\kern-.05em{\sc i\kern-.025em b}\kern-.08em
    T\kern-.1667em\lower.7ex\hbox{E}\kern-.125emX}}

\begin{document}
  \title{G4Seg: Generation for Inexact Segmentation Refinement with Diffusion Models}

\author{Tianjiao Zhang\textsuperscript{1,*}, Fei Zhang\textsuperscript{1,3,*},  Jiangchao Yao\textsuperscript{1,4,\Letter},Ya Zhang\textsuperscript{2,4}, Yanfeng Wang\textsuperscript{2,4,\Letter} \\
  \textsuperscript{1} CMIC, Shanghai Jiao Tong University,
  \textsuperscript{2} School of  Artificial Intelligence, Shanghai Jiao Tong University,\\
  \textsuperscript{3} Shanghai Innovation Institute, 
   \textsuperscript{4} Shanghai Artificial Intelligence Laboratory \\
    \texttt{\{xiaoeyuztj, ferenas, Sunarker, ya\_zhang, wangyanfeng622\}@sjtu.edu.cn}, \\
}

\maketitle

\begin{abstract}
 This paper considers the problem of utilizing a large-scale text-to-image diffusion model to tackle the challenging Inexact Segmentation (IS) task. Unlike traditional approaches that rely heavily on discriminative-model-based paradigms or dense visual representations derived from internal attention mechanisms, our method focuses on the intrinsic generative priors in Stable Diffusion~(SD). Specifically, we exploit the pattern discrepancies between original images and mask-conditional generated images to facilitate a coarse-to-fine segmentation refinement by establishing a semantic correspondence alignment and updating the foreground probability. Comprehensive quantitative and qualitative experiments validate the effectiveness and superiority of our plug-and-play design, underscoring the potential of leveraging generation discrepancies to model dense representations and encouraging further exploration of generative approaches for solving discriminative tasks.
\end{abstract}

% This paper considers the problem of utilizing Diffusion Models (DMs) to achieve a segmentor supervised merely by the image or text signal. Previous discriminative-model-based pipelines,   our goal is to mine Inexact Segmentation by mining out the  to 
%   Diffusion Models (DMs), as long been investigated in  are currently a cutting-edge generation tools been widely for image generation. While the application of diffusion models in recognition tasks remains relatively underexplored, with existing research concentrating on attention extraction and massive synthetic data generation.
% In this paper, we explore a novel paradigm that capitalizes on the inherent guidance capabilities of diffusion models dealing a discriminative task in a generation manner. Under the principle that "better guidance results in better generation", we utilize the discrepancies between images generated with imperfect mask guidance and the original images to refine the coarse segmentation. Extensive experiments has shown the effectiveness of our method. 

\section{Introduction}\label{sec:intro}
Recent breakthroughs in Diffusion Models (DMs) have empowered the field of visual generation for images~\cite{sdm,zhang2024long} and video~\cite{cho2024sora}, demonstrating their capacity of high-fidelity and diverse content synthesis. Meanwhile, there is a growing interest in unlocking DMs for performing the discriminative task of visual dense recognition~\cite{ma2023diffusionseg,zhou2024image}. However, similar to discriminative-model-based segmentation frameworks~\cite{kirillov2023segment,ma2023attrseg, yang2024multi}, these DM-based methods rely heavily on large-scale pixel-level training datasets, which require costly and labor-intensive labeling efforts. To relieve this, this paper explores the potential of DMs in tackling the Inexact Segmentation (IS) problem, a more challenging task that achieves segmentation using only text or image-level class labels, essentially merging two existing settings: Text-Supervised Semantic Segmentation (TSSS)~\cite{ zhang2024uncovering, chen2024probabilistic} and Weakly-Supervised Semantic Segmentation (WSSS)~\cite{ahn2018learning, zhang2021complementary, zhang2022exploiting,mai2023exploit,li2023monte, li2023ddaug, liu2024audio,zhang2024exploiting}.    
\footnote{*~Equal contribution.~\Letter~Corresponding author. }
\footnote{This work is supported by the National Key R\&D Program of China (No. 2022ZD0160703),  National Natural Science Foundation of China (No. 62306178), and STCSM (No. 22DZ2229005), 111 plan (No. BP0719010).}

% \begin{figure}[!th]
%     \centering
%     \includegraphics[width=0.98\linewidth]{figure/teaser_g4.pdf}
%     \vspace{-2mm}
%     \caption{Illustration of our motivation. Previous DM-based methods, similar to discriminative-model-based pipelines, aim to mine out the cross-attention representation for generating the segmentation map. On the contrary,  our proposed training-free method exploits the underlying semantic discrepancy during the generation process, explicitly leading to a coarse-to-fine refinement mechanism.}
%     \vspace{-2mm}
%     \label{fig_teaser}
% \end{figure}

One line of current DM-based IS research is dedicated to excavating and refining the image-text cross-attention map embedded in the noise predictor network~\cite{DiffSegmenter}. 
Specifically, these methods leveraged the object-shape-characterized self-attention module to refine the cross-attention map, yielding a segmentation mask for the query object.
Another line of research focuses on treating a diffusion process as a self-supervised denoising task and employing a diffusion model as a general feature extractor~\cite{odise}. In these studies, diffusion models serve as attention-guiding feature extractors, \textit{indirectly} assisting segmentation tasks. In contrast, research on using generative paradigms to \emph{directly} optimize segmentation remains unexplored, leaving the fundamental generative ability of large-scale pretrained diffusion models underutilized.

In this paper, we delve into the generative nature of pretrained diffusion models to refine a coarse segmentation mask from inexact segmentation.
Specially, we are inspired by cases that GPTs can generate responses closer to the alternative answers under certain prompts to solve discriminative tasks without any extra training.
For visual diffusion models, 
better condition guidance similarly results in a smaller discrepancy between the generated and initial images. Under such an implication, we can use the discrepancy to obtain feedback to improve the condition itself.
Prior work, DiffusionClassifier~\cite{li2023diffusion}, has proved that using a correct text prompt leads to a better denoising result for a specific image, indicating better category classification. We incorporate this spirit into IS, a more challenging discriminative grounding task without pixel-level supervision. A new framework, termed as \textbf{G4Seg}, is proposed, 
which leverages diffusion-based generation with coarse segmentation mask injection and the semantic discrepancy between the generated and initial image 
\begin{figure}[t]
\includegraphics[width=\linewidth]{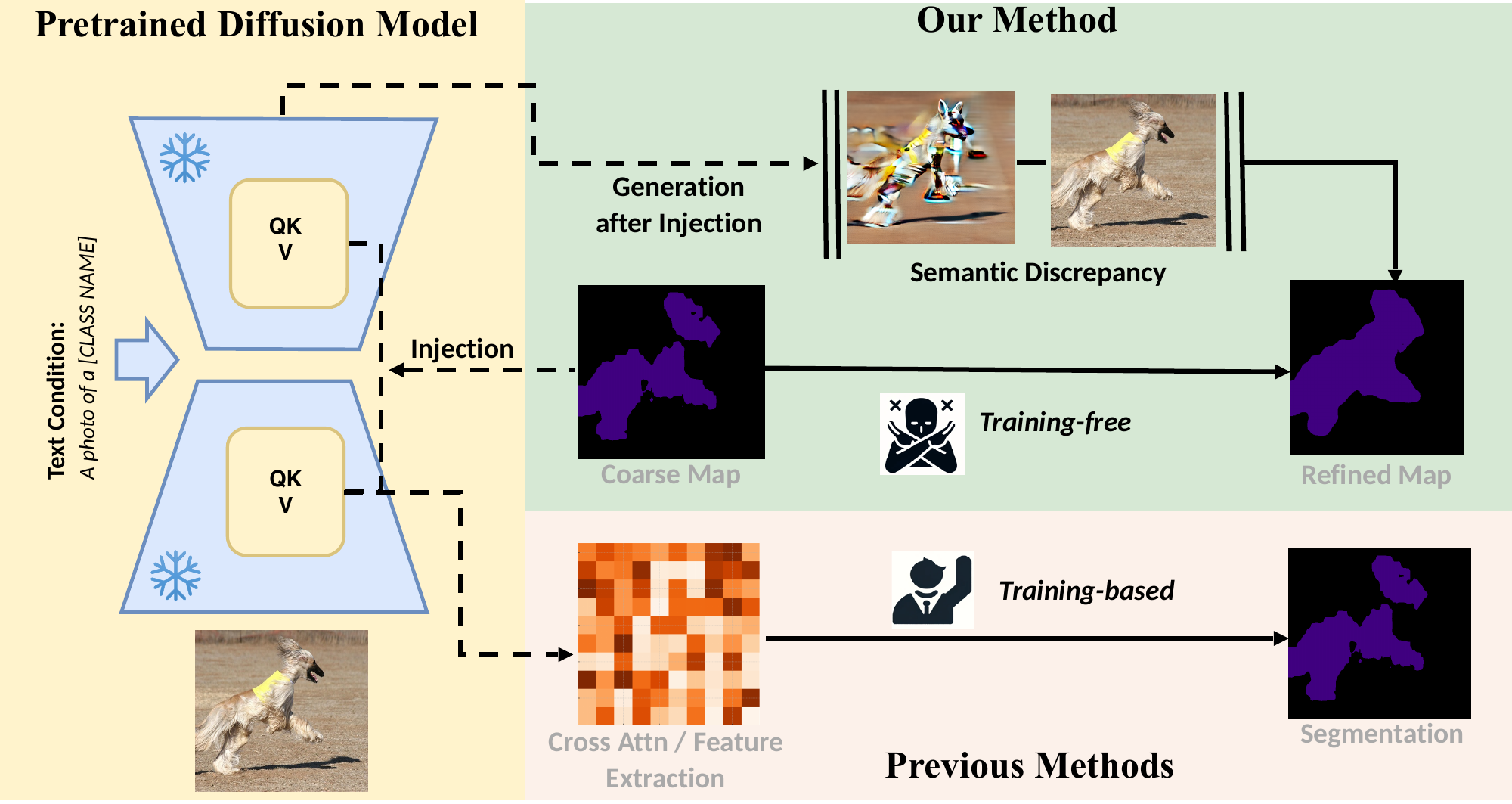}
    \caption{The comparison illustration. Previous DM-based methods mine out features or the cross-attention for generating the segmentation map with training. Our training-free method exploits the underlying semantic discrepancy after intervening the generation process to improve the segmentation map.}
    \label{fig_teaser}

\end{figure}
(as shown in Fig.~\ref{fig_teaser}). It is worth noting that G4Seg is an inference-only framework involving a large-scale pre-trained diffusion model without any extra training or fine-tuning.

Technically, to achieve refinement of the original mask in a generative manner, the image to be segmented should first be inverted into latent noise space or added with noise at a suitable time step. 
Then, the image is reconstructed with the condition, which includes the text prompt and the inexact mask. Under the imperfect mask, the generated image shows some discrepancy from the initial image. By means of the pixel-wise Hausdorff distance as a discrepancy metric, a semantic correspondence alignment methodology is designed for a better inexact segmented mask refinement. 

Our contributions can be summarized as follows: 
\begin{itemize}[leftmargin=20pt]
    \item Different from the popular discriminative-based segmentation paradigm and previous DM-based training methods, we propose a novel training-free framework in a generation manner for inexact segmentation refinement empowered by the condition capacity of pretrained diffusion models.
    \item We are among the first attempts to leverage the discrepancy between original and generated images to refine the coarse mask by technically establishing a principled alignment to build the correspondence and updating the foreground probability of each pixel with its paired pixel.
    %updating the pixel confidence by a probabilistic pair mixing. 
    \item Our framework has achieved a consistent performance gain in both open-vocabulary and weakly supervised segmentation tasks on top of current state-of-the-art methods.
    The methods shed light on using generative models to solve discriminative tasks without training.
\end{itemize}

% Our method is based on a basic evidence that segmentation mask with better quality could guide a better generation. According to the diffusion generation process, our framework is composed of three stages: generation with image-specific inversion feature, dense mask injection and free generation after injection. After obtaining a image generated with coarse mask injection, then we use the both generated and initial image to drive the mask towards a better quality.

\section{Method}
\subsection{Problem Formulation}\label{sec:pf}
Suppose we have an image $\mathcal{I}$ together with its coarse mask $\mathcal{S}_{c}$. It is worth noting that obtaining an inexact coarse segmentation mask $\mathcal{S}_{c}$ is simple and low-cost, achievable through methods like cross-attention extraction~\cite{DiffSegmenter} or by utilizing models such as CLIP~\cite{lin2023clip}.
Towards our goal, we expect to use a pretrained diffusion model $\mathcal{M}$ to first obtain a generated image $\mathcal{I}_g = \mathcal{M}(\mathcal{I}_n; \mathcal{T}, \mathcal{S}_{c})$, where $\mathcal{I}_n$ is the reversed embedding of $\mathcal{I}$ in the noise space and $\mathcal{T}$ is the text prompt. 
Then, by carefully comparing $\mathcal{I}$ and $\mathcal{I}_g$, we will get a mask with better quality $\mathcal{S}_{r} = \Phi(\mathcal{I}_g, \mathcal{I}, \mathcal{S}_{c})$ using the algorithm $\Phi$.

\subsection{Preliminary}
The visual diffusion model works by progressively adding noise to the image in the forward process and then using a deep network to recover the initial image from the pure noise in the backward process.
In the forward process, the clean image $x_0$ is added with Gaussian noise scaled by a specific timestep $t$: $0 \leq t \leq T$, obtaining a noisy sample $x_t = \sqrt{\alpha_t} x_0 + \sqrt{1-\alpha_t} \epsilon$,
where $\alpha_t$ is the pre-defined noise schedules, and $\epsilon \sim \mathcal{N}(0, I)$.
Then a deep learning network $\epsilon_\theta(x_t,t)$ is trained to predict the noise $\epsilon$ from $x_t$:
\begin{equation}
     \mathbb{E}_{t \sim \mathcal{U}(0,T),\epsilon \sim \mathcal{N}(0,I)}[||\epsilon - \epsilon_{\theta}(x_t,t,y)||^2_2,
    \label{Eq:noise}
\end{equation}
where $y$ is the condition. 
With a pre-trained diffusion model, a clean image can be generated from Gaussian noise $p(x_T) \sim \mathcal{N}(0, I)$ step by step by $x_{t-1} = \frac{1}{\sqrt{1-\beta_t}}(x_t-\frac{\beta_t}{\sqrt{1-\alpha_t}}\epsilon_\theta(x_t,t,y)) $.
This can be divided into two substeps. 
The first is to predict the original image $x_0$ (termed as $\tilde{x}_0$ to distinguish from $x_0$) using the current $x_t$ and the model prediction $\epsilon_\theta(x_t,t,y)$:
\begin{equation}
    \tilde{x}_0 = f_\theta(x_t,t;y) = \frac{x_t - \sqrt{1-\alpha_t}\epsilon_\theta(x_t,t,y)}{\sqrt{\alpha_t}}.
    \label{Eq:pred_x0}
\end{equation}
Then, $x_{t-1}$ can be calculated as $x_{t-1} = \sqrt{\alpha_{t-1}}\tilde{x}_0 + \sqrt{1-\alpha_{t-1}-\sigma_t^2}\epsilon_\theta(x_t,t,y) $.
%

% In this work, we employ two types of conditions $y = (y_{\text{txt}}, y_{\text{mask}})$, where the first is the text prompt and the second is the semantic segmentation mask condition.
%
% Then the denoising nature of the diffusion model is employed to refine the segmentation $y_{\text{mask}}$.

\begin{figure*}[!t]
    \centering
\includegraphics[width=16cm]{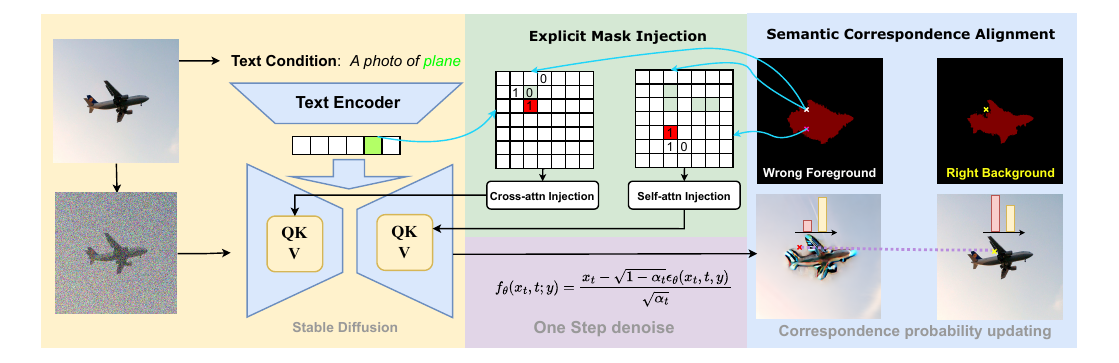}
    \caption{The overall framework of our proposed G4Seg. First, the noisy sample conditioned on the injected coarse mask is fed into the diffusion model to obtain the denoised image. Then, the foreground probability of each pixel is estimated with paired pixels in the semantic correspondence alignment. Finally, the updated segmentation mask is calculated from the pixel foreground probability.
    }
    \label{fig_framework}
\vspace{-3mm}
\end{figure*}

\subsection{G4Seg: A More Efficient Generative Method for Inexact Segmentation}
% In the problem formulation, we generate image with condition on the coarse mask. Due to the heavy generation process of the diffusion model and aiming for a more efficient inference, we simplify the calculation for inverting $\mathcal{I}_n$ into a noise addition and denoising operation. We provide a method based on the full generation with null text inversion~\cite{mokady2023null} in the Appendix.
In Section~\ref{sec:pf}, we generate images conditioned on the coarse mask $S_c$. Normally, we can follow the method of full generation with null text inversion in~\cite{mokady2023null} to achieve near-perfect reconstruction for $\mathcal{I}$. To improve inference efficiency, we simplify the calculation of inverting $\mathcal{I}_n$ into a noise addition and denoising operation. 
For example, given a specific image $x_0$, we first select a candidate timestep $t_s$ and calculate the noisy sample $x_{t_s}$.
Then, we shorten the whole generation process with only one step inference, using Eq.~(\ref{Eq:pred_x0}) to directly get the prediction $\tilde{x}_0$. As for the generation process intervened by the coarse prior $S_c$, we first transform the coarse mask $S_c$ into two masks respectively injected into the cross attention and self-attention of diffusion backbone, which is detailed in Section~\ref{sec:emi}. 
Then, $\tilde{x}_0$ can be calculated under such a mask injection.
Finally, the coarse mask is refined by employing the semantic correspondence alignment between $x_0$ and $\tilde{x}_0$ in Section~\ref{sec:sca}. In the following, we concretely discuss the two critical components of our method.

\subsubsection{Explicit Mask Injection} \label{sec:emi}
% Here, we will introduce the text prompt $ \mathcal{T}$ in the process $y_{mask}^c = \mathcal{T}(S_c)$. 
%
In Stable Diffusion, the textual prompt $\mathcal{T}$ is first tokenized and fed into the CLIP text encoder, forming a textual embedding.
The denoising U-Net then utilizes cross-attention mechanisms with the embedding to leverage textual information for conditioning.
At the same time, self-attention is employed to model the relationships between pixels, which can be leveraged for better generation. 
In this study, we utilize the aforementioned features of cross-attention and self-attention in our diffusion model to inject our prior coarse segmentation mask into the inference process.

Specifically, in the attention layers of diffusion models, the intermediate image feature is first mapped as a query and updated via calculating the attention maps $A \in \mathbb{R}^{q \times k}$ with
$A = \text{softmax}\left(\frac{QK^\top}{\sqrt{d}}\right)$,
where $q$ and $k$ are the lengths of the query $Q$ and the key $K$, which are derived from the context.
The context could either be a text embedding or the image feature itself, noted as cross-attention and self-attention, respectively.
We incorporate the coarse mask as a representation of the ideal correlations between pixels and textual embeddings, integrating it into the generation process of our diffusion model, following the spirit of DenseDiffusion~\cite{kim2023dense}.

For clarity, we flatten the 2D image mask into a 1D signal, facilitating alignment with the 1D textual signal. We assign a superscript to such signals for representation, \textit{e.g.,} $S^{1D}_c$ denotes the coarse mask that is flattened into 1D.
For a coarse mask provided for the category $c$ with the name $T_c$, we prepare the prompt $\mathcal{T}$ as ``\texttt{A photo of {$T_c$}}'' and map it to a textual embedding as the key feature. 

Suppose the index set of $T_c$ in the textual embedding is $\alpha(T_c)$ %\footnote{The index of the class name is represented as a set, as a single long word may correspond to multiple token embeddings, or the class name may consist of two or more words.}.
The injection mask for cross attention $\mathcal{A}_{\text{cross}} \in \mathbb{R}^{q \times k}$ is designed as:
\begin{equation}\label{eq:cross}
    \mathcal{A}_{\text{cross}}(i,j; S_c) =\left\{ \begin{aligned}
        & 1 ~~~\text{if}~~ j \in \alpha(T_c)~~\text{and}~~S_c[i]=1\\ 
        & 0~~~\text{otherwise} 
    \end{aligned}.
    \right.
\end{equation}
$\mathcal{A}_{\text{cross}}(i,j; S_c)$ represents the relation between two types of signals, image, and text, which is set to 1 once the textual embedding and the foreground image token is matched, otherwise 0. Similarly, we can define the injection mask for self-attention $\mathcal{A}_{\text{self}} \in \mathbb{R}^{q \times q}$. However, as the self-attention performs between image tokens, we can compute the mask on the internal $S_c$, which is formulated as follows
\begin{equation}\label{eq:self}
\mathcal{A}_{\text{self}}(i,j; S_c) =\left\{ \begin{aligned}
        & 1 ~~~\text{if}~~ ~S_c[i] = ~S_c[j] =1 \\ 
        & 0~~~\text{otherwise}
    \end{aligned}.
    \right.
\end{equation}
%
% The attention prior mask indicates that the correlation value for any two foreground image tokens will be 1; otherwise, it is 0.
Given two injection masks $\mathcal{A}_{\text{self}}$ and $\mathcal{A}_{\text{cross}}$, we can respectively intervene in the computation of the cross attention and the self-attention in image generation with the pretrained diffusion model
\begin{equation}\label{eq:injection}
A'_{\text{cross/self}} = \text{softmax} (\frac{QK^\top + \alpha\mathcal{A}_\text{cross/self}}{\sqrt{d}}),
\end{equation}
where the $\alpha$ is the injection weight. By incorporating the intervention of the coarse mask $S_c$ by Eq.~\ref{eq:cross}, Eq.~\ref{eq:self} and Eq.~\ref{eq:injection}, the image generation result $\tilde{x}_0$ is affected.
In the next section, we will illustrate how to employ the gap between the reconstructed image $\tilde{x}_0$ and the original image $x_0$. 

\subsubsection{Semantic Correspondence Alignment}\label{sec:sca}
With explicit mask injection, we have obtained a model that can generate images conditioned on the coarse mask. In other words, without loss of generality, we have a generative model $p(x|S)$, where $S$ is a given mask and $x$ is the target image. However, in a segmentation task, we actually want to find $\max_S p(S|x)$ given an image $x$. Intrinsically, they can be connected by using Bayes' Law as below:
\begin{equation}
    \max_S p(S|x) = \max_S \frac{p(x|S)p(S)}{p(x)} \Leftrightarrow \max_S p(x|S),
\end{equation}
since $p(x)$ and $p(S)$ should be constant for a specific $x$. Here, the segmentation task can be treated as a conditional generation problem, which fits our intuition that \textit{with more accurate mask condition, the probability of generating $x$ is more likely to be maximized.} Following this spirit, we assume that $p(x|S)$ follows a distribution that is inversely related to $d(x,\tilde{x}(S))$, namely $p(x|S)\propto -d(x,\tilde{x}(S))$, where $\tilde{x}(S)$ denotes the corresponding generation conditioned on the mask $S$, and $d$ represents an image-wise distance measure. Consequently, the problem reduces to $\min_S d(x,\tilde{x}(S))$. In this study, we realize the image-level distance measure by means of the pixel-level Hausdorff distance~\cite{huttenlocher1993comparing}, denoted as $d_\text{Haus}(\cdot,\cdot)$, which provides us the inspiration of transforming the optimization into a semantic correspondence alignment based on image discrepancy, $\max_S p(x|S) \xrightarrow{\text{reduce}} \min_S d_\text{Haus}(x,\tilde{x}(S))$, then the $S$ could be updated with:
\begin{equation}\label{eq:haus}
 S[j] \leftarrow S[j] + \gamma ~ \frac{\partial D(x[{\delta_j}], \tilde{x}(S)[j])  }{\partial S[j]},
\end{equation}
where $\tilde{x}(S)[j]$ denotes the $j$th pixel in the generated image $ \tilde{x}(S)$ and $\delta_j$ denotes the index of the corresponding pixel in the original image $x$ that requires to be searched. $D$ is a pixel metric based on the semantic gap between two pixels. The detailed deduction can be found in Supplementary.
For a specific category as foreground, if we treat $S[j]$ as the foreground probability, we can interestingly observe that $S[j]$ is updated based on the discrepancy~($D$ in the Equation, which denotes as a semantic gap) between the probability of the pixel in the original and generated images.

Despite the potential insight inherent in Eq.~\ref{eq:haus}, it is intractable due to the discrete operation implemented on the coarse mask $S$ in Eq.~\ref{eq:cross} and Eq.~\ref{eq:self}. However, we can follow its spirit to build a semantic correspondence alignment to achieve a similar goal. That is: 1) we first find the optimal pixel alignment $\delta_j$ as in Eq.~\ref{eq:haus}; 2) and then we use a simple linear mixing between paired pixels in the generated and initial images to approximate the segmentation (foreground) mask updating direction. 
% With coarse mask $S_c$, we firstly establish a correspondence mapping between $(\tilde{x}(S_c)$ and initial $x_0$ following dift~\cite{dift}, which is also in a training-free manner. 
%
For the first step, we use a predefined feature extractor $F(\cdot)$ to embed the generated and original images into the feature space, denoted as $F(\tilde{x}(S))$ and $F(x)$. 
For the $j$th pixel in the generated image, the corresponding point $\delta_j$ can be searched via:
\begin{equation}
    \delta_j = \arg\min_{j'} \mathcal{D}(F(x)[j'], F(\tilde{x}(S))[j]),
\end{equation}
where $\mathcal{D}(\cdot,\cdot)$ denotes the cosine similarity metric defined in the feature space for semantic correspondence alignment.
For a specific pixel, we obtain the pixel-wise feature from the image feature and then search for the pixel in the original image that has the smallest distance.
Then, for the second step, we can estimate the foreground probability $S^\star$ at position $j$ as follows,
\begin{equation}
    S^\star[j] = \beta S[j] + (1-\beta) S[\delta_j],
    \label{eq:mix}
\end{equation}
where $\beta$ is the mixing coefficient.
Finally, with the refined foreground probability $S^\star[\cdot]$ of each pixel, we obtain the refined segmentation mask.

\section{Experiments}~\label{sec_Experiments}
\vspace{-4mm}
\subsection{Implementation Details}~\label{sec_Experiments_Implementation Details}
\noindent\textbf{Datasets and Evaluation Metric.} Following~\cite{lin2023clip}, we evaluate G4Seg on three benchmarks, i.e., PASCAL VOC12 (20 foreground classes)~\cite{voc12}, PASCAL Context (59 foreground classes)~\cite{context}, and  MS COCO Object 2014 datasets~\cite{coco} (80 foreground classes). 
All of these datasets contain 1 extra background class. During the inference, only the image-level (class) label is used to generate the mask. The mean Intersection-over-Union (mIoU) is adopted as the evaluation metric (\%). For each specific class, We treat all other segments as background and update the current segment logit independently.

\subsection{Inexact Segmentation Performance}~\label{sec_Zero_shot_Segmentation_Performance}
\begin{table}[t]

\centering
\caption{Comparison with TSSS methods.}
\label{tab_TSSS_perforamnce}
\resizebox{0.5\textwidth}{!}{
\begin{tabular}{lccc}
\toprule[1.4pt]
Methods & VOC12  & Context & COCO \\
\midrule
\multicolumn{4}{l}{\emph{Training-based}} \\
ViL-Seg~\cite{liu2022open}  & 34.4 & 16.3 & 16.4 \\
TCL~\cite{cha2022learning}  & 51.2 & 24.3 & 30.4 \\
GroupViT~\cite{xu2022groupvit}  & 52.3 & 22.4 & 20.9 \\
ViewCo~\cite{viewco}  & 52.4 & 23.0 & 23.5 \\
SegCLIP~\cite{luo2022segclip}  & 52.6 & 24.7 & 26.5 \\
PGSeg~\cite{zhang2024uncovering}  & 53.2 & 23.8 & 28.7 \\
OVSegmentor~\cite{xu2023learning} & 53.8 & 20.4 & 25.1 \\
\rowcolor{lightgray}
G4Seg{\footnotesize +\emph{GroupViT}} & 53.4\textcolor{tgreen}{\scriptsize\textbf{+1.1}} & 23.9\textcolor{tgreen}{\scriptsize\textbf{+1.5}} & 22.1\textcolor{tgreen}{\scriptsize\textbf{+1.2}}   \\
\midrule
\midrule
\multicolumn{4}{l}{\emph{Training-free}} \\
% ReCo~\cite{shin2022reco}  & 25.1 & 19.9 & 15.7 \\
MaskCLIP~\cite{zhou2022extract}  & 38.8 & 23.6 & 20.1 \\
SCLIP~\cite{wang2023sclip}  & 59.1 & 30.4 & 30.5 \\
DiffSegmenter~\cite{DiffSegmenter} & 60.1 & 27.5 & 37.9 \\
\rowcolor{lightgray}
G4Seg{\footnotesize +\emph{SCLIP}} & 59.8\textcolor{tgreen}{\scriptsize+\textbf{0.7}} & \textbf{31.3}\textcolor{tgreen}{\scriptsize+\textbf{0.9}} & 30.9\textcolor{tgreen}{\scriptsize+\textbf{0.4}}   \\
\rowcolor{lightgray}
G4Seg{\footnotesize +\emph{DiffSegmenter}} & \textbf{60.6}\textcolor{tgreen}{\scriptsize+\textbf{0.5}} & 
28.1\textcolor{tgreen}{\scriptsize+\textbf{0.6}}
& \textbf{38.5}\textcolor{tgreen}{\scriptsize+\textbf{0.6}}   \\
\bottomrule[1.4pt]
    
\end{tabular}
}
\vspace{-3mm}
% }
% \end{table}
\end{table}
\noindent\textbf{Performance on TSSS.} Here we first evaluate the performance of our method in TSSS. Table~\ref{tab_TSSS_perforamnce} lists the mIoU of 10 state-of-the-art (SOTA) methods on the validation of PASCAL VOC12, PASCAl Context, and COCO Object. 
Notably, these methods are categorized into two splits, i.e., \emph{training-based} and \emph{training-free}, and we implement our on-top-of method based on 3 methods (1 training-based + 2 training-free). 

\begin{figure}[t!]
% \vspace{-2mm}
\centering
\includegraphics[width=\linewidth]{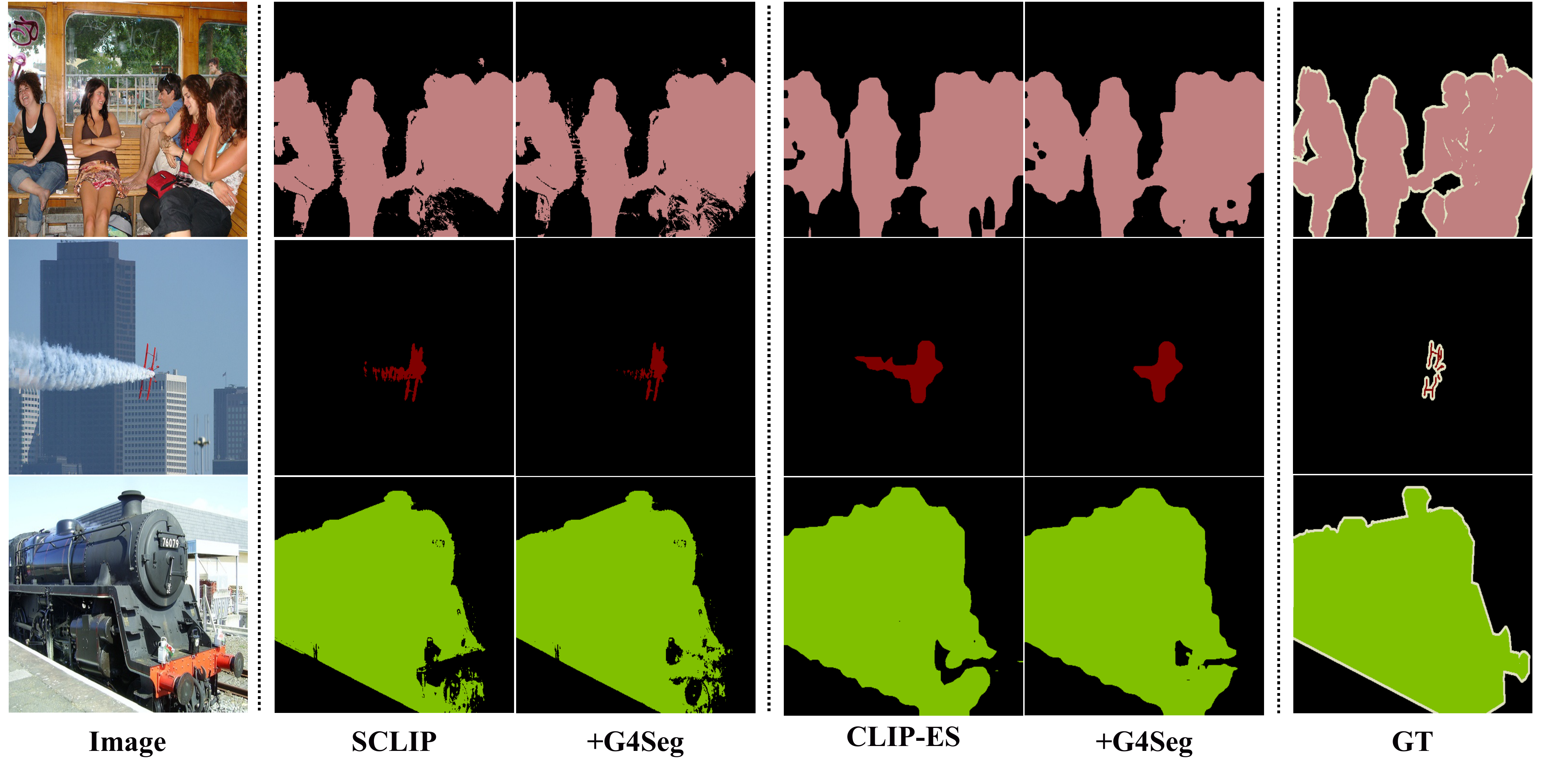}
\vspace{-3mm}
    \caption{Qualitative results on PASCAL VOC12. Compared with the baseline, G4Seg could further segment the object in a more complete and delicate way.  }
    \label{fig_seg}
\vspace{-3mm}
\end{figure}

Note that our method does not involve any training process. As shown in this table, it is clear that our method, regardless of the training paradigm, could achieve an overall improvement compared to all the adopted baseline methods, with an average elation of \textbf{0.77}\%, \textbf{1.00}\%, and \textbf{0.73}\% across these three benchmarks. Additionally, with such prominent improvement, our method yields new SOTA performance against all methods in TSSS. Fig~\ref{fig_seg} shows some illustrative samples for a visualized comparison, validating the effectiveness and superiority of our method in open-domain segmentation refinement. 

\begin{table}[t]
% \begin{table}[!htbp]
\centering
\caption{Comparison with WSSS methods on VOC12 \emph{train}. The mask is generated from Seed refined with Post-processing  (Post.) approaches. * denotes that ~\cite{zhu2023weaktr} adopts a designed self-training strategy.
}
\label{tab_WSSS_perforamnce}
\resizebox{0.5\textwidth}{!}{
\begin{tabular}{lcccc}
\toprule[1.4pt]
Methods & Post. & Seed & Mask \\
\midrule
CAM~\cite{ahn2018learning}  & dCRF & 48.0 & 52.4 \\
IRN~\cite{ahn2019weakly}  & RW+dCRF & 48.5 & 63.5 \\
SEAM~\cite{wang2020self}  & RW+dCRF & 55.4 & 63.6 \\
MCTformer~\cite{xu2022multi}  & RW+dCRF & 61.7 & 69.1 \\
ViT-PCM~\cite{rossetti2022max}  & dCRF & 67.7 & 71.4 \\
ToCo~\cite{ru2023token}  & - & 73.6 & 73.6 \\
WeakTr~\cite{zhu2023weaktr}  &  Self-Training* & 66.2 & 76.5 \\

CLIP-ES~\cite{lin2023clip}  & dCRF & 70.8 & 74.9 \\
\midrule
G4Seg{\footnotesize +\emph{CAM}} & dCRF  & 50.8\textcolor{tgreen}{\scriptsize+\textbf{2.8}} & 54.2\textcolor{tgreen}{\scriptsize+\textbf{1.8}}   \\
% \rowcolor{lightgray}
G4Seg{\footnotesize +\emph{CLIP-ES}} & dCRF  &  \textbf{72.0\textcolor{tgreen}{\scriptsize\textbf{+1.2}}} & 

\textbf{75.4\textcolor{tgreen}{\scriptsize \textbf{+0.5}}}   \\
\bottomrule[1.4pt]
    
\end{tabular}
}
% }
\vspace{-4mm}
% \end{table}
\end{table}

\noindent\textbf{Performance on WSSS.} Here we compare our methods with a line of works in WSSS. As downstream training is required, WSSS evaluates the model's ability to segment task-specific objects. In this way, to evaluate the effectiveness of our method in task-specific learning, Table~\ref{tab_WSSS_perforamnce} reports the performance of our method in comparison with 8 prevailing WSSS frameworks. Here we would like to emphasize that two post-processing refining mechanisms are commonly utilized in WSSS, i.e., RW~\cite{ahn2019weakly} and dCRF~\cite{chen2017deeplab}, which helps \emph{refine the coarse Seed into the fine-grained Mask}. As shown in Table~\ref{tab_WSSS_perforamnce}, our method based on two WSSS frameworks achieves an overall consistent improvement compared to the adopted baselines, leading to an average accuracy increase of \textbf{2.0}\% (\textbf{1.2}\%) on Seed (Mask). 

These experimental results further validate the versatility of our method in domain-specific segmentation. In this way, our approach with CLIP-ES yields a new SOTA performance in WSSS, further demonstrating the excellence of our training-free method in zero-shot IS. Fig~\ref{fig_seg} showcases some illustrative samples that are produced from the baselines and our methods.

% Compared with TSSS, WSSS aims to achieve a closed-domain segmentor \emph{simply with image-level class labels}

\subsection{Ablation Studies}~\label{sec_Ablation_Studies}
In this Section, unless specifically specified, we use the Seed of G4Seg with CLIP-ES to implement all ablation studies on PASCAL VOC12 in detail, which mainly contains the effectiveness of the modules in G4Seg, the influence of time step, and some illustrative visualized results.

\begin{table}[ht]
\vspace{-3mm}
\centering
\caption{Ablation studies on the modules in G4Seg.}
\label{tab_eachmodule}
\vspace{-3mm}
\resizebox{1.0\linewidth}{!}{
\setlength{\tabcolsep}{1mm}{
\begin{tabular}{c c c c c  | c}
\toprule
Baseline & EMI  & SCA  & CF-[0.2,0.6] & CF-[0.1,0.7]   & mIoU (\%)\\
\midrule 
\textcolor{teal}{ \CheckmarkBold}  &             &      &    &     &  70.8\\
\textcolor{teal}{ \CheckmarkBold}  &    \textcolor{teal}{ \CheckmarkBold}        &      &    &     &  71.3\textcolor{tgreen}{\scriptsize +0.5} \\
\textcolor{teal}{ \CheckmarkBold}  &  \textcolor{teal}{ \CheckmarkBold}          &  \textcolor{teal}{ \CheckmarkBold}     &    &     &  71.7\textcolor{tgreen}{\scriptsize +0.9} \\
\textcolor{teal}{ \CheckmarkBold}  &   \textcolor{teal}{ \CheckmarkBold}          &   \textcolor{teal}{ \CheckmarkBold}     &  \textcolor{teal}{ \CheckmarkBold}  &     &  \textbf{72.0\textcolor{tgreen}{\scriptsize+1.2}} \\
\textcolor{teal}{ \CheckmarkBold} &   \textcolor{teal}{ \CheckmarkBold}           &   \textcolor{teal}{ \CheckmarkBold}      &    &  \textcolor{teal}{ \CheckmarkBold}    &  71.6\textcolor{tgreen}{\scriptsize +0.8} \\
\bottomrule
\end{tabular}
}
}
\vspace{-2mm}
% \end{table}
\end{table}
\noindent\textbf{Effectiveness of Individual Module.} Table~\ref{tab_eachmodule} presents the effectiveness of each individual module in G4Seg. As shown in this table, adding EMI could explicitly bring a certain elation (\textbf{+0.5\%}) compared with the baseline, indicating the benefits of mask injection during the denoising stage. Additionally, further improvements achieved through SCA (\textbf{+0.9\%}) demonstrate that establishing a correspondence between the mask-injected image and the original image can emphasize the importance of key matching points for fine-grained segmentation. We also propose the CF strategy to further improve the performance of SCA by matching and modifying the most uncertain points. Consequently, it is observed a proper setting of the filtering range could yield the boosting of G4Seg (\textbf{+1.2\%}), achieving a final \textbf{72.0\%} performance together with all modules.

\begin{figure}[ht]
\vspace{-3mm}
\centering
    \includegraphics[width=0.9\linewidth]{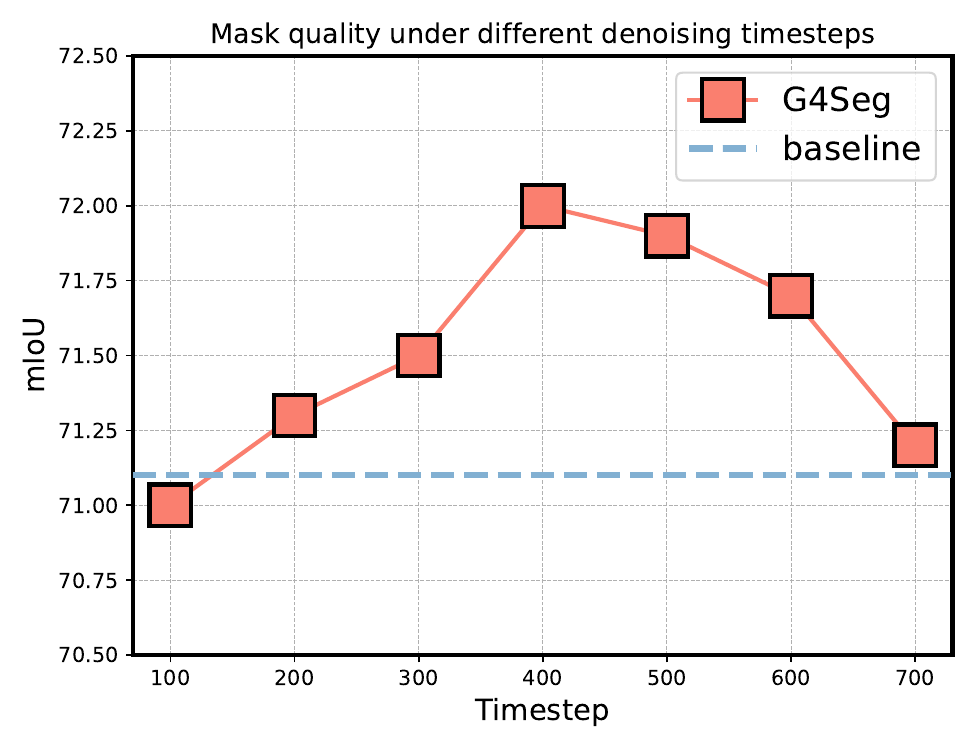}
    \caption{The mask quality with different timestep noise.}
    \label{fig_den_timesteps}
\vspace{-2mm}
\end{figure}

% \begin{wraptable}{r}{0.65\textwidth}
% \centering
% \vspace{-4mm}
% \caption{The mask quality under different denoising timestep.}
% \vspace{-2.5mm}
% \resizebox{1.0\linewidth}{!}{
% % \setlength{\tabcolsep}{1mm}{
% \begin{tabular}{ c  |c|c|c|c|c|c|c}
% \toprule[1pt]
%   Timestep & 100 & 200 & 300  & 400 & 500 & 600 & 700  \\
%   \hline
%  \vspace{-3mm}\\
%     mIoU  & 71.0 & 71.3 & 71.5 & \textbf{72.0} & 71.9 & 71.7 &71.2 \\
% \bottomrule[1pt]
% \end{tabular}
% }
% % }
% \label{tab_timestep}
% \vspace{-3mm}
% \end{wraptable}
\noindent\textbf{Different Timesteps.}  G4Seg adopts the fixed noising-denoising step for the generated image. To investigate the impact of the denoising timestep, we conduct our method by setting different timesteps obtained from \{100, 200, 300, 400, 500,600,700\}. As shown in Fig.~\ref{fig_den_timesteps}, it can be observed our method is robust to the timestep due to a small performance fluctuation. The best performance is achieved at step 400, and then the larger/smaller timestep could yield a performance decrease.    

\begin{table}[ht]
\centering
% \vspace{2mm}
\caption{The influence of Null-text Inversion (NtI). The unit of speed is second(s) for processing one sample.}
\vspace{-2.5mm}
\resizebox{1.0\linewidth}{!}{
\begin{tabular}{ c  |c|c |c|c|c}
\toprule[1.4pt]
  Method & \multicolumn{2}{c}{G4Seg+GroupViT / + NtI} & & \multicolumn{2}{c}{G4Seg+CLIP-ES / + NtI} \\
  \midrule
  mIoU & \multicolumn{2}{c}{54.0 / 54.2} & & 
  \multicolumn{2}{c}{72.0 / 72.1} \\ 

  Speed(s) & \multicolumn{2}{c}{+1.2/5.5}\ & & 
  \multicolumn{2}{c}{+1.1/5.2} \\ 
 % \vspace{-3mm}\\
 %    mIoU  & 71.0 & 71.3 & 71.5 & \textbf{72.00} & 71.9 & 71.7 &71.2 \\
\bottomrule[1.4pt]
\end{tabular}

}
% }
\label{tab_NtI}
\end{table}

\noindent\textbf{Involvement of Null-text Inversion.} G4Seg utilizes the difference between the generated and original image to help refine the mask. Due to the single-step noising-denoising process, it is hard to flawlessly reconstruct the original image. Intuitively, here we explore \emph{whether better reconstruction could bring more explicit improvement.} To this end, we introduce Null-text Inversion (NtI) for our method, achieving near-perfect reconstruction by finding the corresponding initial noise during the inversion. Table~\ref{tab_NtI} reports the performance and the inference speed comparison between our method and NtI-involved paradigm. Interestingly, the involvement of NtI simply showcases the marginal improvement as expected. 

% considering change this to the appendix

\section{Limitations and Future Work}
Though promising performance is achieved by our method, such a DM-based method inevitably meets the comparably slow-mask-inference issue due to the sampling denoising process. Besides, since SD is simply trained on natural images, our training-free method may not be applicable in some non-natural image domains, such as medical and agricultural imaging. Finally, due to resource limitations, we do not implement the latest SD version (SD-XL, SD3, Flux), and additional tuning on our method is also not achieved, both of which shall lead to better segmentation performance. 

\section{Time and Memory Efficiency}
Our G4Seg could be implemented on simply 1 RTX 3090 GPU, generating 1 mask at a time and occupying 15GB. Since our method simply requires a direct single-step noising-denoising process to the original image, G4Seg could finish the inference of all 1449 images in VOC12 validation images within 1.5 hours (3 seconds per image), leading to a reasonable level of computational efficiency. Note that adopting multiple GPUs or multiprocessing could further speed up the inference process. In fact, we implement our G4Seg with 4 3090 GPUs. In this way, the inference time is reduced to about 18 minutes.

\section{Implementation Details}
\noindent\textbf{Inference Settings} Our model is fully based on Stable Diffusion~2-1\cite{sdm}, which is trained on LAION~\cite{schuhmann2022laion}. In our experiment, all images are resized to (512,~512). All experiments are merely conducted on 1 RTX 3090 GPU equipped with 24 GB of memory without any extra training. Our method, working in an on-top-of manner, follows a \emph{refine-after-generate} paradigm: generating the masks from the selected mask-free baseline first and then refining them via our proposed method without additional training. For the mask generation process, we strictly follow the settings in the selected baselines. For explicit mask injection, our parameter follows the DenseDiffusion~\cite{kim2023dense} and the added noise step is 400.
For the semantic correspondence alignment, the feature extractor we adopt is a CLIP image encoder to better distinguish between generated and initial image. The pixel correspondence mixing coefficient $\beta$ is set to 0.8 for open-vocabulary segmentation and 0.9 for weakly-supervised semantic segmentation.

\section{A Hausdorff distance view of correspondence alignment}
\label{sec:haus}
Here we illustrate our method in a more theoretical view. Suppose we have a mask $S$ conditioned generation model, which could estimate $p(x|S)$.Then, we want to inverse this process with $p(S|x)$ which denotes that given a $x$ the $S$ distribution should be estimated. So in a segmentation task, we want to estimate:
\begin{equation}
    \max_S p(S|x),
\end{equation}
where $x$ denotes specific samples. Owing to the law of condition probability:
$$
p(S|x) = \frac{p(x|S) p(S)}{p(x)}.
$$
For the given $x$ and suppose all the segmentation masks share the same probability, we omit the $p(x)$ and $p(S)$ terms. Then the final result is equivalent as the:
\begin{equation}
    \max_S p(x|S)~~~~ \text{with specific}~x.
\end{equation} 
where indicates our institution, \textbf{with accurate mask condition, the probability of generating $x$ is maximized.} This is truly our basic stone.

Here we make further assumption, owing to the Gaussian essence of diffusion generation, the $p(x|S)$ could be estimated by:
\begin{equation}
    p(x|S) \propto \exp(- d(x,\tilde{x}(S))^2),
\end{equation}
where the $\tilde{x}(S)$ denotes the generating $\tilde{x}$ based on S. Then the problem is equivalent to $ \min_{S} d(x,\tilde{x}(S)) $. The problem becomes, finding a more appropriate mask, then minimum the gap between the mask-conditioned generation and initial image.

Then, we based on this update the $S$ with stochastic gradient descent:
\begin{equation}
    S = S + \gamma ~ \frac{\partial d(x,\tilde{x}(S)) }{\partial S},
    \label{lab:deri}
\end{equation}
where $\gamma$ denotes the step size. Then we consider a Hausdorff distance between two images $(A,B)$ with pixel-wise $(a,b)$ distance:
\begin{equation}
    H(A,B)= \sup_{a \in A}~~\inf_{b \in B}~~D(a,b),
\end{equation}
where $D$ denotes the pixel-wise distance to distinguish from the image-wise distance $d$. Here we consider the initial image and conditioned generated image,
\begin{equation}
H(\tilde{x}(S),x)= \sup_{\tilde{x}(S)_j \in \tilde{x}(S)}~\inf_{x[i] \in x}~~D(x[i], \tilde{x}(S)[j]),
\end{equation}
where the $[i,j]$ indicates the i'th and j'th pixel of the initial and generated image. If we carefully look at the $ \inf_{x[i] \in x}~D(x[i], \tilde{x}(S)[j])$, \textbf{the term indicates the correspondence pixel among all $x[i]$s in $x$ with minimum distance towards $\tilde{x}(S)[j]$}. Here we consider an equivalent formation substituting superb with summation.
\begin{equation}
H'(\tilde{x}(S),x)= \sum_{\tilde{x}(S)[j] \in \tilde{x}(S)}~D(x[\delta_j], \tilde{x}(S)[j]),
\label{lab:distance}
\end{equation}
where $x_{\delta_j}$ denotes the correspondence point with $\tilde{x}(S)[j]$. With specific $j$'th pixel $\tilde{x}(S)[j]$, we substitute Eq.~{\ref{lab:distance}} into Eq.~{\ref{lab:deri}}, then we obtaining:
\begin{equation}
     S[j] \leftarrow S[j] + \gamma ~ \frac{\partial D(x[\delta_j], \tilde{x}(S)[j])  }{\partial S[j]},
\end{equation}
where this could be treated as mask optimization and updating process. 

\section{Visualization for G4Seg and Null text inversion}
\begin{figure*}[tbp]
\vspace{-2mm}
\centering
\includegraphics[width=0.92\textwidth]{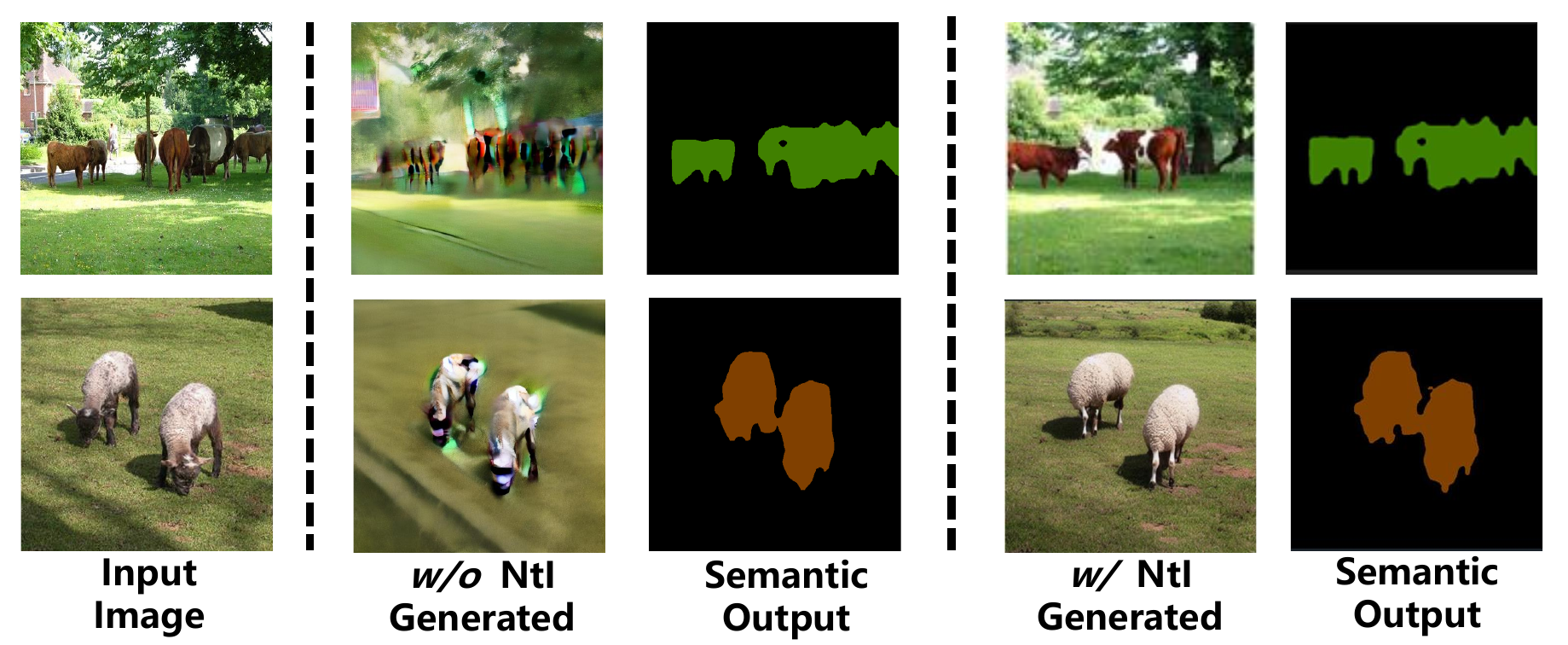}
\vspace{-2mm}
    \caption{Visualized comparison between G4Seg \emph{w/} NtI and \emph{w/o} NtI. }
    \label{fig_nti}
\vspace{-5mm}
\label{fig:ntic}
\end{figure*}
The results could be found in Figure.~\ref{fig:ntic}.

\begin{figure}[ht]
    \includegraphics[width=1.0\linewidth]{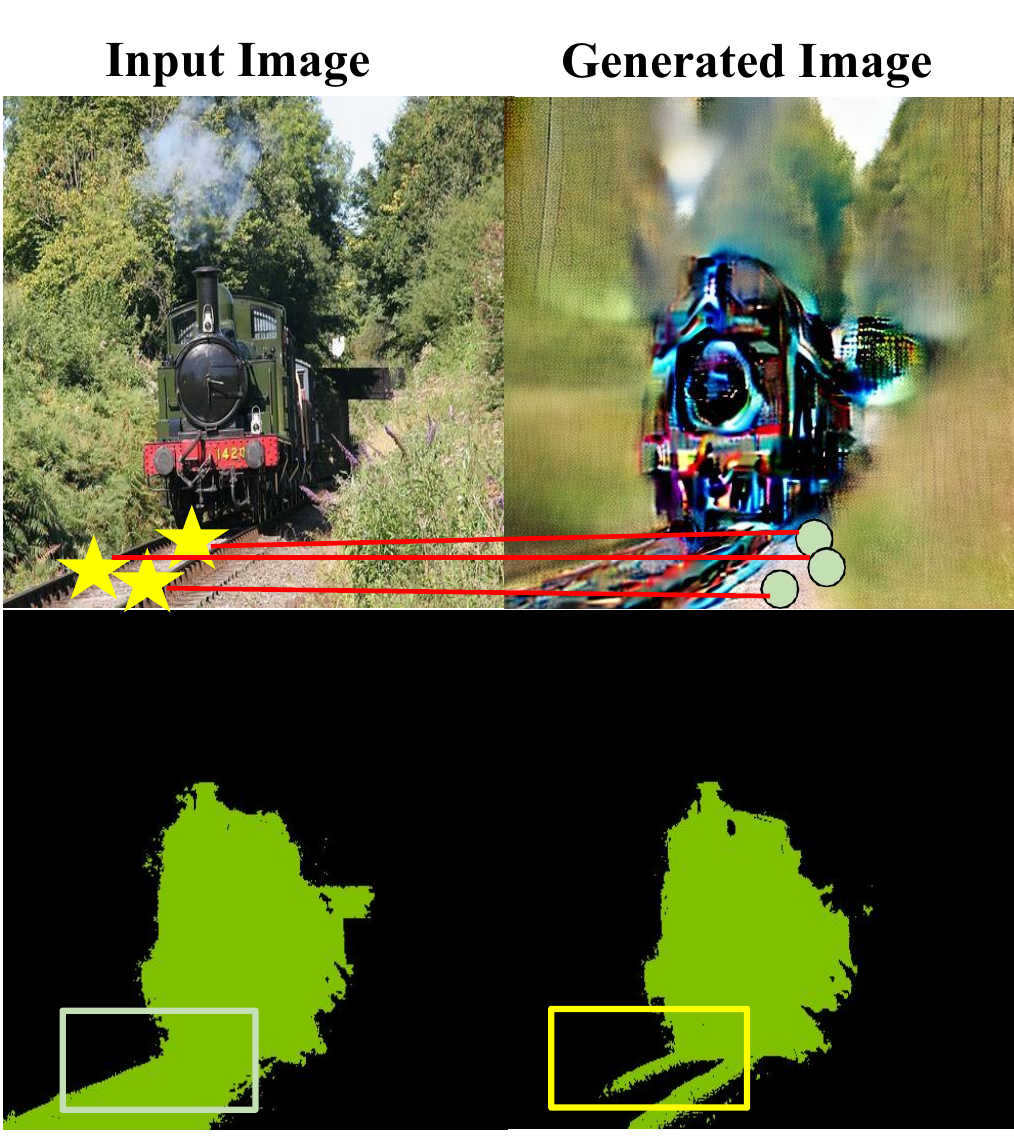}
    \vspace{-5mm}
    \caption{Visualization analysis on SCA. SCA is able to correct the wrongly segmented pixels based on the generated-original image discrepancy.}
    \label{fig_correspondence}

\end{figure}
\section{Correspondence Analysis.}
Correspondence learning has been widely proven to be efficient in many previous works~\cite{zhang2025slice,zhang2024tracking}.
Our proposed SCA explicitly refines the mask by building the feature-level semantic correspondence. 
Figure~\ref{fig_correspondence} presents the visualized correspondence matching map. SCA builds a one-to-one mapping between the original (stars) and mask-injected generated images (circles). The matched pixel from the generated image (marked by the small circles) reflects the same semantic content as the original image (marked by the stars). However, with the coarse mask injection, the generated image shall have wrongly-recognized regions for the query object. 
Specifically, there is a generated semantic of ``train'' for the railroad in the generated image, 
which is the result of the over-segmented coarse mask (marked by the green box). With the help of correspondence alignment, the mis-segmented pixel is corrected to embrace the appropriate object regions, relieving the over-segmented regions. In this way, we observe that the incorporation of correspondence helps improve the boundary regions. Such fine-grained refinement further validates the effectiveness of G4Seg, demonstrating the rationality of adopting generation discrepancy in segmentation which is consistent with the discussion in the main paper.

\section{Adaptive adjustment for over/under-segmented area}
\label{app:aaa}
\begin{figure}[h]
\centering
\includegraphics[width=0.98\linewidth]{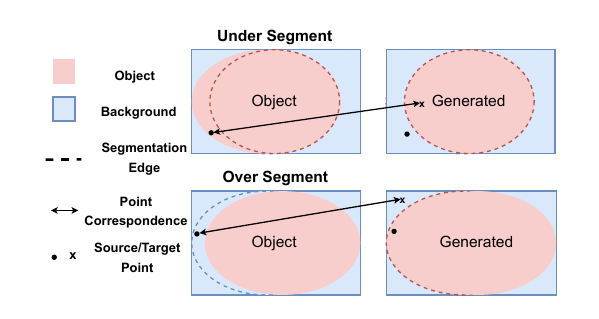}
\vspace{-1.5mm}
    \caption{the linear mixing of foreground probability of paired pixels could adaptively adjust the over-/under segmented area. }
    \label{fig_sca_th}
\end{figure}

Different types of errors can introduce various impacts on the generated results. We make a more detailed discussion on the Figure~\ref{fig_sca_th}. We divided these errors into two categories and discussed the points in each area:

\subsection{Over-segmented}

\textbf{Definition}: segmenting some background as foreground. 

\textbf{Phenomenon}: The generated area tends to expand, incorporating the semantic of the object into areas that were originally background, as shown in the first row of Figure~\ref{fig_sca}.

\textbf{Segmentation refining}: The corresponding point moves to the exterior with a lower probability. Then the mixing in 
\begin{equation}
    S^\star[j] = \beta S[j] + (1-\beta) S[\delta_j],
    \label{eq:mix}
\end{equation}
would lead to a lower foreground probability, the point is more likely to be recognized as background correctly. 

\subsection{Under-segmented}

\textbf{Definition}: segmenting some foreground as background. 

\begin{figure*}[t]
% \vspace{-2mm}
\centering
\includegraphics[width=0.98\textwidth]{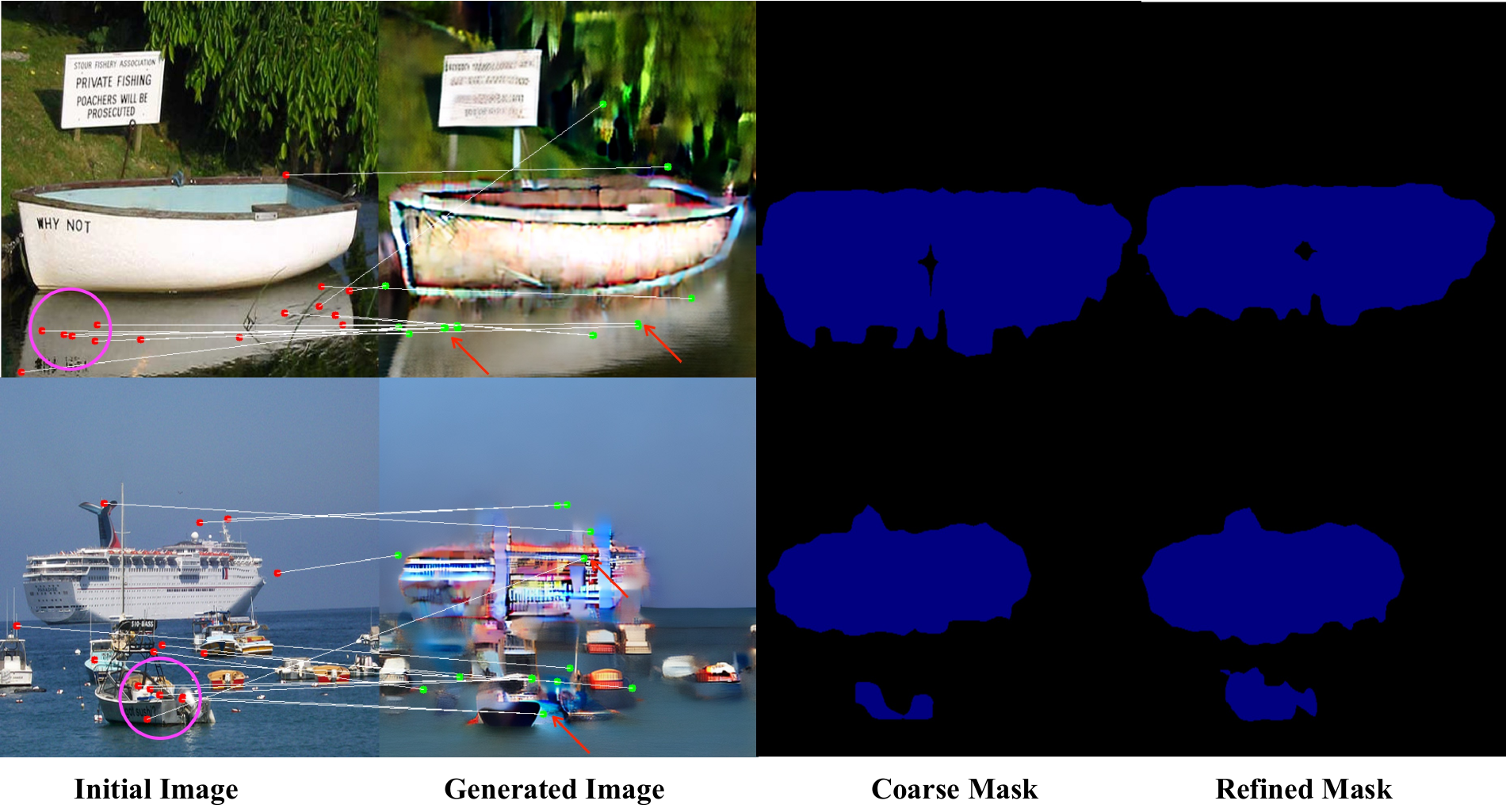}
\vspace{-1.5mm}
    \caption{Visualization of correspondence and segmentation refining. Random 15 points are selected for visualization. In the ships in the second row, the coarse segmentation reveals that the middle part of the smaller ship is missing. The semantics of pixels in the middle part are eroded by the background in generated image. The points under-segmented in red circles are mapped to the edges of the ship and the hull of the larger ship with higher target probability. }
    \label{fig_sca}
\end{figure*}

\textbf{Phenomenon}: The generated object tends to shrink, converting areas originally belonging to the object into the background, as shown in the second row of Figure~\ref{fig_sca}.

\textbf{Segmentation refining}: The corresponding point moves to the interior of the object with increasing foreground probability. Then after the points probability mixing, the point is more likely classified as foreground correctly.

In summary, our method generates corresponding defective images based on the flaws in the existing segmented mask. The mixed probability is then adaptively adjusted according to different scenarios. A visualization result can be found in Figure~\ref{fig_sca}.

\section{On top of fully/semi-supervised methods}
\textbf{Fully/semi-supervised open-vocabulary semantic segmentation}
To better evaluate our methods, we build G4Seg on top of some fully/semi-supervised open-vocabulary semantic segmentation methods: 
\begin{itemize}
  
   \item OVAM~\cite{marcos2024open}: OVAM uses manually annotated masks of generated images to update token embeddings, which are then used to generate more images and corrected cross-attention-based pseudo masks. 
   \item
    DeOP~\cite{Han2023ZeroShotSS} DeOP is inherently a fully supervised method trained with precisely annotated pixel labels.
\end{itemize}

\begin{table}[!ht]
    \centering
    \begin{tabular}{c|cc}
    \toprule
        Methods & VOC & Context \\
        \midrule
        OVAM & 61.2 & 28.3 \\ 
        +G4Seg & 62.1(+0.9)	 &	28.9(+0.6) \\ 
        DeOP & 91.7 & 48.8 \\ 
        +G4Seg & 92.1 (+0.4) & 49.3(+0.5) \\
    \bottomrule
    \end{tabular}
\end{table}

\textbf{Fully supervised closed setting}
Our method relies on a pre-trained diffusion model and allows for sample-wise segmentation improvement by providing the image and its corresponding coarse mask. For closed-set semantic segmentation, we conduct our method on ADE20K using three fully supervised segmentation approaches: SegFormer~\cite{xie2021segformer}, Mask2Former~\cite{cheng2022masked}, and ODISE~\cite{odise}, which cover semantic segmentation and panoptic segmentation.
\begin{table}[!t]
    \centering
    \caption{Closed set fully supervised semantic segmentation}
    \begin{tabular}{c|c}
    \toprule
        Methods & mIoU/PQ \\ \midrule
        SegFormer (B1) & 42.2 \\ 
        +G4Seg & 42.9(+0.7) \\ 
        Mask2Former(R50) & 47.2 \\ 
        +G4Seg & 47.8(+0.6) \\ 
        ODISE(panoptic) & 22.4 \\ 
        +G4Seg & 23.0(+0.6) \\ \bottomrule
    \end{tabular}
\end{table}

\textbf{Fully supervised cross-domain semantic segmentation}

We evaluate the performance of our method on a cross-domain setting and adopt a baseline~\cite{wei2023disentangle} for nighttime semantic segmentation on NightCity-fine~\cite{tan2021night}.

\begin{table}[!h]
    \centering
        \caption{Cross domain fully supervised semantic segmentation}
    \begin{tabular}{c|c}
    \toprule
        Methods & mIoU \\ \midrule
        DP~\cite{wei2023disentangle} & 64.0 \\ 
        DP+G4Seg & 64.5(+0.5) \\ \bottomrule
    \end{tabular}
\end{table}

\section{Sensitivity assessment on coarse mask quality before refinement}
Ideally, our method does not rely on the initial mask quality. To show how sensitive the proposed method relying on the initial segmentation quality, since our approach is sample-wise, we performed stratification based on different quality levels of coarse segmentation and then calculated the mean IoU improvement for samples with different levels for the VOC dataset:
\begin{table}[!ht]\scriptsize
    \centering
    \caption{Performance versus initial quality}
    \begin{tabular}{c|ccc}
    \toprule
       Initial Quality(IoU range) & 0-40 & 40-80 & 80-100 \\ \midrule
        \# samples(ratio) & 56(3.4\%) & 679(47.3\%) & 237(49.3\%) \\ 
        Avg G4Seg Gain & +0.2 & +1.9 & +1.1 \\ 
         Avg Controlnet~\cite{zhang2023adding} Gain & +0.75 & +4.2 & +4.1 \\
        Avg CascadedPSP~\cite{cheng2020cascadepsp} Gain & +0.2 & +1.5 & +1.0 \\
        \bottomrule
    \end{tabular}
\end{table}
When the initial mask quality is poor, the improvement of our method is also limited. The improvement from our method is most significant for initial IoU values between 40 and 80. This indicates that our approach is particularly effective when the initial segmentation is already of reasonable quality. When the initial segmentation is already nearly perfect(80-100), the improvement from our method becomes limited due to the bottleneck caused by errors inherent in the mask injection process.

 For initial coarse mask for relatively low performance,  all three methods (including one fully supervised method) have limited improvements compared to higher initial segmentation. Since we have introduced a mask conditioning method which is based on cross-attention and self-attention, is the performance bottleneck. These attention-based generation methods do not perform well in a mask-conditioned generation. If we adopt a stronger mask conditioning method, such as Contronet, the performance would significantly improve

\section{Results with other mask injection method}

The overall pipeline of G4Seg is firstly obtaining a mask $S$ conditioned generative models $p(x|S)$ then updating the mask using the generative result with coarse mask. In first step, for serving the in-exact nature, we only use the attention perturbation in diffusion backbone avoiding involving exact pixel-level annotation.

Pursuing a better result with permission to use a pixel-level annotation, we could involving a more stronger mask injection method, Controlnet~\cite{zhang2023adding}. The ControlNet consists of approximately half of a diffusion backbone and functions as a feature extractor that can accept arbitrary signals (such as segmentation masks) as input. The extracted features are then integrated into the diffusion backbone to control the generative output, $\epsilon(x_t,t,S)$. For images-annotation pairs($x_0$ and S), then the controlnet is trained with:
$$
\mathcal{L}_{cn} = E_{\epsilon \sim N(0,I)}||\epsilon-\epsilon(x_t,t,S)||_2^2
$$

For our implementations, we use the pretrained segmentation conditioned model provided by~\cite{segcontrolnet} which is then fine-tuned on the corresponding training set with the nearest palette defined by~\cite{segcontrolnet}.

With an improved pipeline, the performance would significantly improve, as shown in the following table:
\begin{table}[!ht]\small
\caption{Performance with different injection strategy}
    \centering
\begin{tabular}{c|ccc}
\toprule
Method(G4Seg)                    & CLIP-ES$_\text{VOC}$ & SCLIP$_\text{COCO}$ & SCLIP$_\text{Context}$ \\
\midrule
+EMI(Attn Injection) & 72.0        & 30.9       & 31.3          \\
+EMI(Controlnet~\cite{zhang2023adding})     & 74.1        & 33.1       & 33.8 \\        
\bottomrule
\end{tabular}

\end{table}

\section{Comparison with related works}
\label{app:comp}
\textbf{On-top-of}. Our method, as a plug-and-play framework, can be simply and efficiently integrated into various existing segmentation modules to enhance the performance online with the current single sample.

\textbf{Generative content with generated-original bias}. Some work such as VPD~\cite{zhao2023unleashing} and ODISE~\cite{odise} use pretrained diffusion model as feature extractor with a self-supervised denoising loss. While another line of research, such as OVDiff~\cite{ovdiff} and Freeda~\cite{freeda}, merely utilize the content directly generated by diffusion models for target prototype retrieval. In our work, we explore the discrepancy between the generative content and the initial image to refine the discriminative result.

\textbf{Discriminative assistance}. Some diffusion-based training-free segmentation works such as Freeda~\cite{freeda} and OVDiff~\cite{ovdiff} employ pre-trained discriminative models such as DINO~\cite{dino} as assistance, while the performance of the framework heavily depends on these discriminative models.

\textbf{Cross attention initialization}. Most works employ the attention between text and image as a segmentation prior to the diffusion model, such as DatsetDiffusion~\cite{gen_data_3} and DiffSegmentor~\cite{DiffSegmenter}. The most significant difference between our work and others is that the attention mechanism we used is EMI as injecting the coarse mask prior to the generation pipeline. The EMI part could be substituted without attention with another more advanced mask-injecting module.

\section{G4Seg implemented with different diffusion version}

We have compared the results with SD1.5, SD2.1, SDXL and LCM. SD1.5, SD2.1, LCM, and SDXL share largely similar U-Net backbone architectures, incorporating cross-attention and self-attention layers. Consequently, the EMI step is executed in a nearly identical manner across these models. Then after the generation, the SCA step remains the same.
\begin{table}[!ht]
    \centering
    \caption{Performance versus different version of pretrained diffusion model}
    \begin{tabular}{c|c}
    \toprule
        Diffusion Version & mIoU \\ \midrule
        SD1.5 & 71.8 \\ 
        SD2.1 & 72.0 \\ 
        SDXL & 72.0 \\ 
        LCM & 72.1 \\ \bottomrule
    \end{tabular}
\end{table}

\section{Comparison with training-free diffusion segmentation methods}
As the table shows, the DiffSegmentor only relies on the attention mechanism in the diffusion backbone as the clue to the target mask, which does not fully excavate the generation prior to the diffusion model. The OVDiff and Freeda utilize many generated images with a specific class and obtaining the discriminative prototype of the class, where the prototype is retrieved based on the region of interest from cross/Self-attention aggregation. Due to the strong external discriminative assistance, it is challenging to determine whether the generative capacity of the diffusion model contributes to the performance. Our method aims to fully exploit the generative prior for a discriminative task, specifically inexact segmentation, by adopting a GPT-like approach to solve the discriminative task in a generative manner without any extra assistance. 

\begin{table*}[!ht]
    \centering
    \caption{Comparision with training-free diffusion segmentation methods }
    \begin{tabular}{c|ccc}
    \toprule
        Training-free Methods & Generative Content & Gen->Seg & discriminative assitance \\ \midrule
        OVDiff[1] & Class conidtioned images & Cross/Self attention & discriminative feature prototype \\
        Freeda[2] & Class conidtioned images & Cross/Self attention & discriminative feature prototype \\ 
        DiffSegmentor[3] & None & Cross/Self attention & None \\ 
        G4Seg & Mask conidtioned images & Semantic correspondence updating & None \\ \bottomrule
    \end{tabular}
\end{table*}

\section{More Results on Comparison with Other Mask Refinement Methods}
{We have also conducted a comparison between other mask refinement methods on VOC and Context datasets with SCLIP and MaskCLIP.
\begin{table*}[!ht]
    \centering
    \caption{Comparison with different refining methods}
    \begin{tabular}{c|cccc}
    \toprule
        Methods & SCLIP VOC & SCLIP Context & MaskCLIP VOC & MaskCLIP Context \\ \midrule
        Baseline & 59.1 & 30.4 & 38.8 & 23.6 \\ 
        +G4Seg & 59.8(+0.7) & 31.3(+0.9) & 39.4(+0.6) & 24.1(+0.5) \\
        +SegRefiner & 59.3(+0.2) & 30.7(+0.3) & 39.1(+0.3) & 23.9(+0.3) \\ 
        +CascadePSP & 59.5(+0.4) & 30.9(+0.5) & 39.2(+0.4) & 23.8(+0.2) \\ 
        +Densecrf & 60.9(+1.8) & 31.2(+0.8) & 39.9(+1.1) & 24.2(+0.6) \\ 
        +G4Seg + CascadePSP & 60.1(+1.0) & 31.6(+1.2) & 39.5(+0.7) & 24.3(+0.7) \\ 
        +G4Seg+Densecrf & \textbf{62.1(+3.0)} & \textbf{32.0(+1.6)} & \textbf{40.1(+1.3)} & \textbf{24.6(+1.0)} \\ \bottomrule
    \end{tabular}
\end{table*}
\section{More Visualized Results}\label{ap_sec_More_Visualized_Results}
\subsection{Dataset Details}
\noindent\textbf{Datasets.} We evaluate our G4Seg on 3 prevalent benchmarks, which are PASCAL VOC12 2012~\cite{voc12}, COCO~\cite{coco}, PASCAL Context~\cite{context}. Here is the detailed introduction of these five datasets as follows:
\begin{enumerate}[leftmargin=*,label=\textbf{\textbullet}]
\item \textbf{PASCAL VOC2012~\cite{voc12}:} The PASCAL VOC12 dataset consists of a diverse collection of images spanning 21 different object categories (including one background class), such as a person, car, dog, and chair. The dataset provides annotations for both training and validation sets, with around 1,464 images in the training set and 1,449 images in the validation set. We use the validation set for the downstream evaluation.

\item \textbf{COCO~\cite{coco}:} The COCO Object dataset covers a wide range of 80 object categories, such as cars, bicycles, people, animals, and household items. For semantic segmentation, it has 118,287 training images and 5,000 images for validation. 

\item \textbf{Context~\cite{context}:} The dataset contains a diverse set of images taken from various scenes, including indoor and outdoor environments. It covers 59 common object classes, such as a person, car, bicycle, and tree, as well as 60 additional stuff classes, including sky, road, grass, and water. It has 118,287 training images and 5,000 images for validation. Here we merely consider the object dataset part and use the validation set. 

\end{enumerate}

\section{Visualization of reconstruction results from different timesteps}
Figure~\ref{fig_timestep} shows one illustrative sample generated with different timesteps. Clearly, a timestep that is too small, representing a minor perturbation to the original image, would reasonably yield insufficient knowledge injection. Conversely, a timestep that is too large results in a substantial visual discrepancy between the generated and original images, leading to invalid correspondence matching.      

\begin{figure*}[!t]
\vspace{-2mm}
\centering
\includegraphics[width=0.95\textwidth]{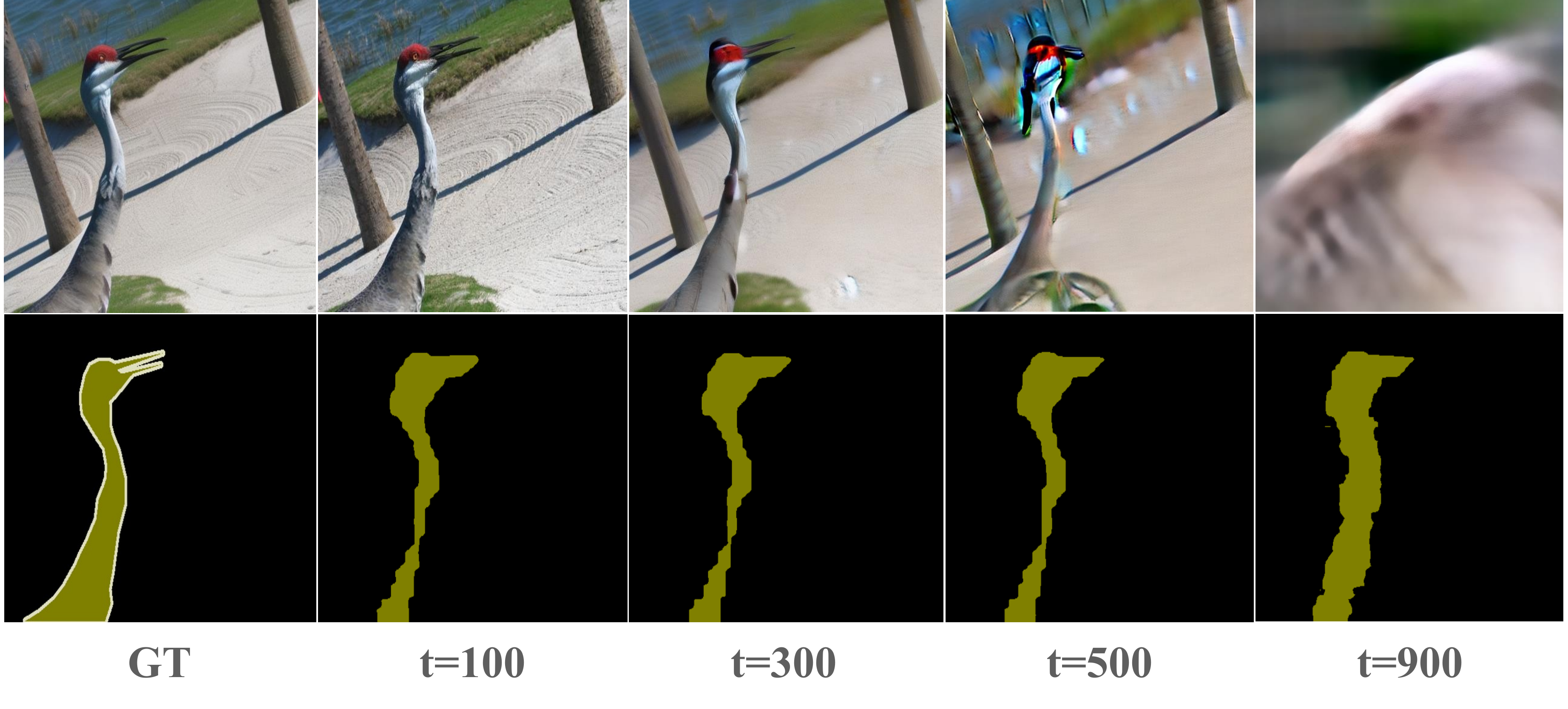}
\vspace{-2mm}
    \caption{Visualized analysis of G4Seg under different denoising timesteps. }
    \label{fig_timestep}
\vspace{-3mm}
\end{figure*}

\subsection{VOC results}
Figure~\ref{fig_voc_ap} presents more results of our G4Seg in VOC12. It is found that our G4Seg shows powerful grouping capability when segmenting the object-centric images. Besides, the generated discrepancy could help segment objects in a compact and dense manner, which means there is less redundancy and noise in objects.

\begin{figure*}
\centering
\includegraphics[width=0.8\textwidth]{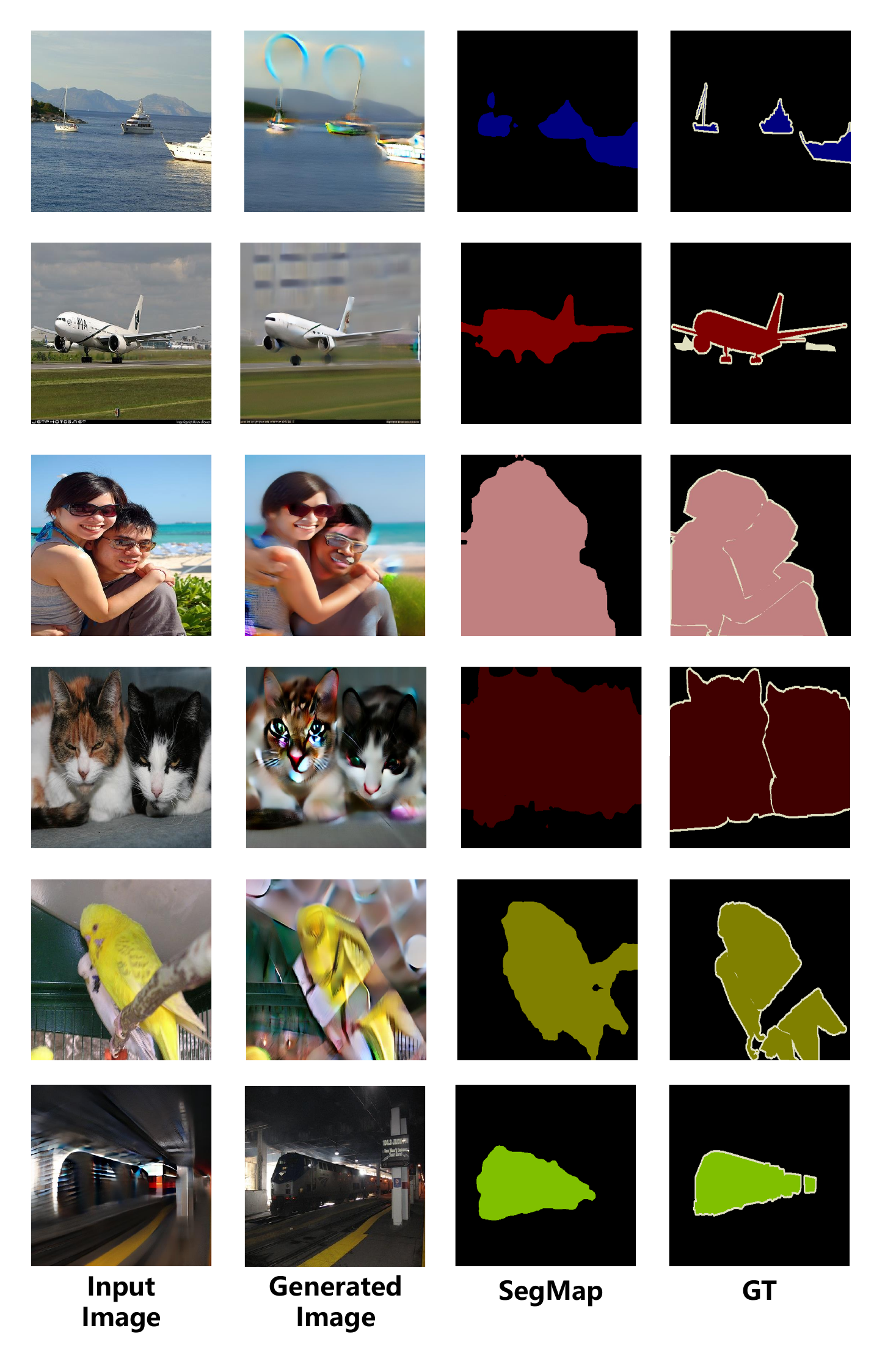}
    \caption{Qualitative results on PASCAL VOC12.}
    \label{fig_voc_ap}
\end{figure*}

\subsection{COCO results}
Figure~\ref{fig_coco_ap} presents some visualized results of COCO Object. Clearly, it has been observed that, compared to GroupViT, our G4Seg is able to perform fine-grained segmentation in the multi-object case. However, G4Seg is unable to provide full areas of object segmentation, revealing the bottleneck of our method.

\begin{figure*}
\centering
\includegraphics[width=0.8\textwidth]{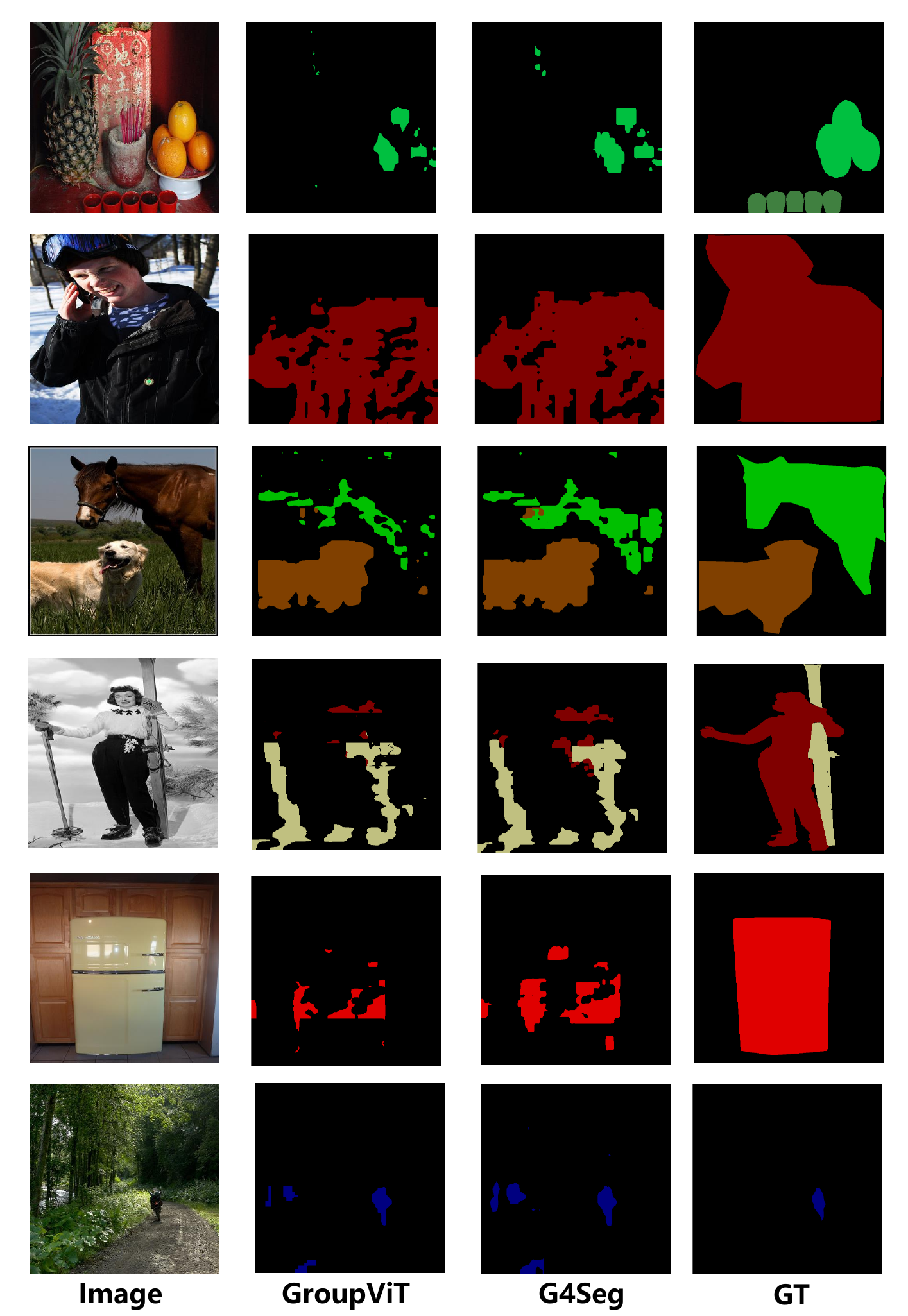}
    \caption{Qualitative results on COCO Object.}
    \label{fig_coco_ap}
\end{figure*}

\subsection{Context results}
Figure~\ref{fig_context_ap} shows several visualized results of Context. A similar improvement could be observed. Besides, G4Seg could enhance the discriminative regions to a large extent in some cases, indicating its effectiveness in multi-object learning.

\begin{figure*}
\centering
\includegraphics[width=0.8\textwidth]{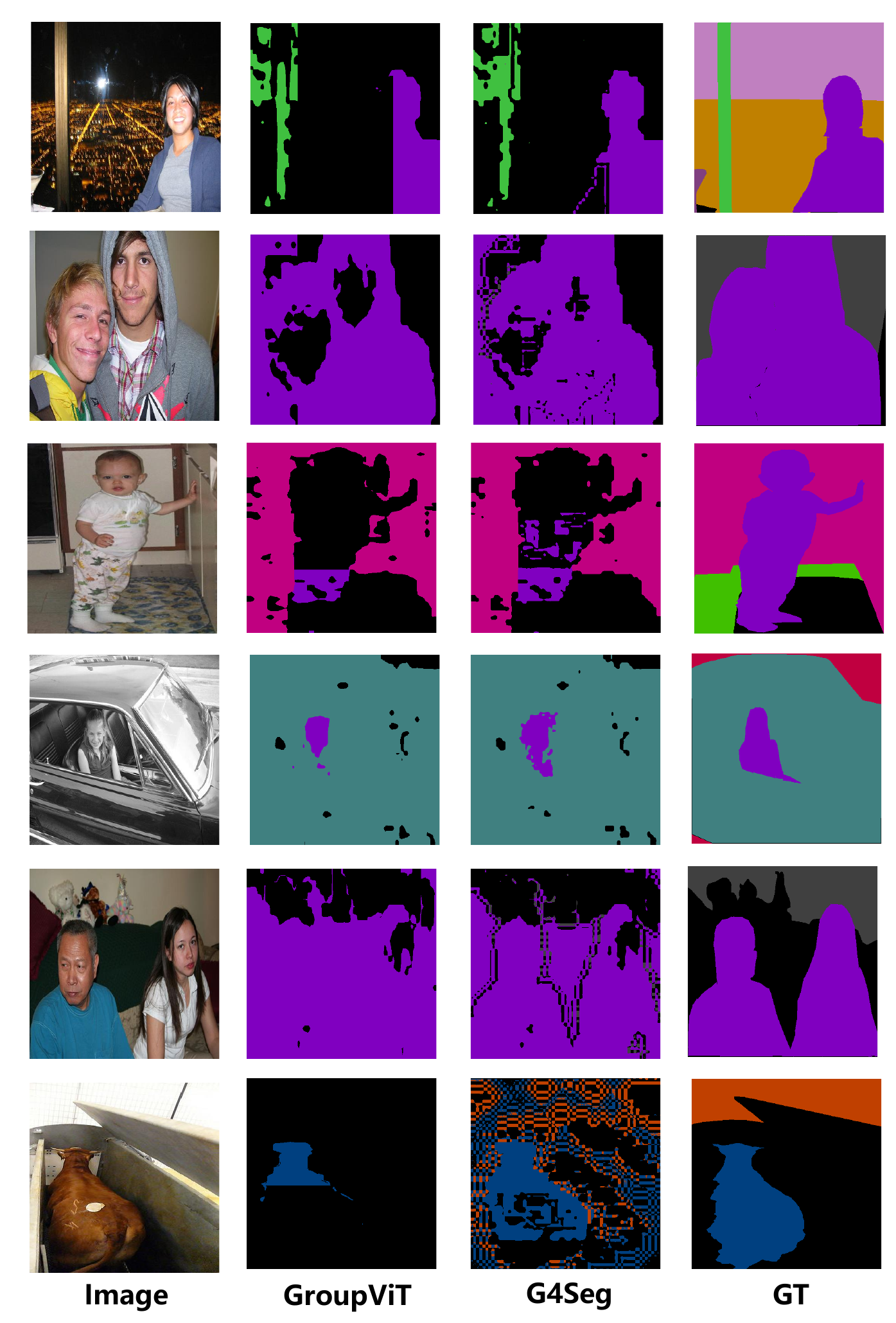}
    \caption{Qualitative results on Context.}
    \label{fig_context_ap}
\end{figure*}
% In all, the utilization of such generation discrepancy, highlighting the generative characteristic of diffusion models, validates its potential in modeling dense representation. Thus, we hope this work could sparkle the exploration of diffusion models in discriminative task, which warrants further exploration in this direction.    

\section{Conclusion}
% This paper explored an intuitive yet feasible training-free solution based on Stable Diffusion (SD), a representative large-scale text-to-image diffusion model, to tackle the challenging vision task of Inexact Segmentation (IS), which aims to achieve segmentation using only texts or image-level labels as minimal supervision. Most existing works, whether utilizing discriminative models or transferring diffusion models, focus on exploiting the visual dense representations inherently arising from the internal attention mechanism. In contrast, this paper delved into the underlying generative priors in SD, specifically the pattern discrepancies between the original and mask-conditioned reconstructed images, to encourage a coarse-to-fine segmentation refinement by progressively aligning the generated and original representations. Furthermore, we proposed establishing pixel-level semantic correspondence between the generated and original patterns, leading to precise corrections towards flawless segmentation for matching points. Through quantitative and qualitative experiments, we demonstrated the effectiveness and superiority of this plug-and-play design. Overall, the utilization of generation discrepancies, which highlight the generative characteristics of diffusion models, validates their potential in modeling dense representations. We hope this work will inspire further exploration of diffusion models in discriminative tasks.

This paper explored an intuitive yet feasible training-free solution based on Stable Diffusion (SD), a representative large-scale text-to-image diffusion model, to tackle the challenging vision task of Inexact Segmentation (IS), which aims at achieving segmentation using merely texts or image-level labels as minimalist supervision. Most SD-based trials, following the discriminative-model-exploited pipelines,  fall into the pure exploitation of the visual dense representations inherently arising from the inner attention mechanism. In contrast, this paper emphasized the underlying generation prior in SD, i.e., the pattern discrepancy between the original and mask-conditioning reconstructed images, to encourage a coarse-to-fine segmentation refinement by progressively aligning the generated-original representations. Furthermore, we proposed establishing the pixel-level semantic correspondence between the generated-original patterns, yielding a delicate correction towards flawless segmentation for the matched point.  Through quantitative and qualitative experiments, we have demonstrated the effectiveness and superiority of this plug-and-play design. Our results highlight the potential of utilizing generation discrepancies to model dense representations in diffusion models. We hope this work inspires further exploration of diffusion models in discriminative tasks.

\bibliographystyle{IEEEbib}
\bibliography{reference}

\end{document}